\documentclass[runningheads]{llncs}
\usepackage[T1]{fontenc}

\usepackage{graphicx}
\usepackage{color}
\usepackage{graphicx,wrapfig,lipsum}
\usepackage{amsmath}
\usepackage{graphicx}
\usepackage{caption}
\usepackage{enumitem}   
\usepackage{amssymb}
\usepackage[boxed, ruled, vlined, linesnumbered]{algorithm2e}

\newcommand{\Dist}[0]{{\rm Dist}}
\newcommand{\Reg}[0]{{\rm Reg}}

\DeclareMathOperator{\E}{E}

\begin{document}
%
\title{MetaMorph: Learning Metamorphic Image Transformation With Appearance Changes}
\titlerunning{MetaMorph: Learning Metamorphic Image Transformation}
\author{Jian Wang  \inst{1}\and
 Jiarui Xing \inst{2} \and
 Jason Druzgal \inst{3} \and
 William M. Wells III \inst{4,5} \and
Miaomiao Zhang\inst{1,2}}

\authorrunning{Wang et al.}
%
\institute{Computer Science, University of Virginia,  VA, USA \and
Electrical and Computer Engineering, University of Virginia,  VA,USA \and
Radiology and Medical Imaging, University of Virginia,  VA,USA \and
Brigham and Women's Hospital, Harvard Medical School, MA, USA
\and Computer Science and Artificial Intelligence Laboratory, MIT, MA, USA 
}

\maketitle              
\begin{abstract}
This paper presents a novel predictive model, MetaMorph, for metamorphic registration of images with appearance changes (i.e., caused by brain tumors). In contrast to previous learning-based registration methods that have little or no control over appearance-changes, our model introduces a new regularization that can effectively suppress the negative effects of appearance changing areas. In particular, we develop a piecewise regularization on the tangent space of diffeomorphic transformations (also known as initial velocity fields) via learned segmentation maps of abnormal regions. The geometric transformation and appearance changes are treated as joint tasks that are mutually beneficial. Our model MetaMorph is more robust and accurate when searching for an optimal registration solution under the guidance of segmentation, which in turn improves the segmentation performance by providing appropriately augmented training labels. We validate MetaMorph on real 3D human brain tumor magnetic resonance imaging (MRI) scans. Experimental results show that our model outperforms the state-of-the-art learning-based registration models. The proposed MetaMorph has great potential in various image-guided clinical interventions, e.g., real-time image-guided navigation systems for tumor removal surgery.
  
\end{abstract}

\section{Introduction}
Deformable image registration is an important tool in a variety of medical image analysis tasks, such as multi-modality image alignment~\cite{maes1997multimodality,chung2002multi,qin2019unsupervised}, statistical analysis for population image studies~\cite{zhang2016low,rao2008hierarchical,wang2021bayesian}, atlas-guided image segmentation or classification~\cite{riklin2010segmentation,wachinger2014atlas,wang2022geo}, and object tracking with anomaly detection~\cite{chen1999anomaly,prastawa2004brain}. In many clinical applications, it is desirable that the estimated transformations are diffeomorphisms (i.e., bijective, smooth, and inverse smooth mappings) because they produce anatomically plausible images~\cite{beg2005computing}. Despite recent achievements in treating the problem of diffeomorphic image registration as a fast learning task, current approaches oftentimes have an assumption that the topology of objects presented in images is intact~\cite{balakrishnan2019voxelmorph,wang2020deepflash,kim2021diffusemorph,chen2022transmorph}. Existing algorithms fail badly in cases where appearance changes occur (e.g., missing data caused by pathology, such as tumors, myocardial scars, multiple sclerosis, and etc.) because they have little to no control over these unknown variables.  

To address this issue, a few algorithms of image metamorphosis have been developed to incorporate the modeling of appearance changes in registration functions~\cite{patenaude2011bayesian,holm2009euler,franccois2021metamorphic,niethammer2011geometric,bone2020learning}. Existing metamorphic image registration methods mainly fall into two categories: (i) exclude appearance changes via manually delineated segmentations of abnormal regions~\cite{patenaude2011bayesian,niethammer2011geometric}, and (ii) treat the appearance changes as unknown variables estimated out from images~\cite{franccois2021metamorphic,bone2020learning}. These approaches either heavily depend on manually segmented labels of 3D volumetric data that are time and labor-consuming, or struggle with balancing between the effects of appearance vs. geometric changes. A recent work~\cite{bone2020learning} has developed a metamorphic autoencoder that  estimates the deformation and appearance variations by decoupling the geometric and appearance representations in latent spaces. However, such a model is highly sensitive to parameter-tuning due to its difficulty in differentiating changes caused by geometric transformations vs. appearances.

In this paper, we develop a novel learning-based model of metamorphic image registration, named as MetaMorph, that provides more robust and accurate registration results in images with appearance changes. In contrast to previous approaches~\cite{patenaude2011bayesian,franccois2021metamorphic,niethammer2011geometric,bone2020learning}, we incorporate a new {\em appearance-aware regularization} in the network loss function that enforces a piecewise constraint on geometric transformation fields. Such a constraint will be learned simultaneously from a jointly optimized segmentation task. In addition, we effectively augment the segmentation labels by utilizing the learned transformations in the training process. This not only substantially improves the segmentation performance, but also reduces the requirement for massive ground truth segmentation labels. The main contributions of our proposed MetaMorph are summarized in three folds: 
\begin{itemize}
    \item To the best of our knowledge, MetaMorph is the first predictive registration algorithm that utilizes jointly learned segmentation maps to model appearance changes.
    \item MetaMorph learns a new appearance-aware regularization that piecewisely constrains the variations of image intensities caused by geometric transformations separately from appearance changes. 
    \item The joint learning scheme of MetaMorph maximizes the mutual benefits of metamorphic image registration and segmentation. 
\end{itemize}

To demonstrate the effectiveness of our model, we validate MetaMorph on real 3D human brain tumor MRIs. Experimental results show that MetaMorph outperforms the state-of-the-art learning-based registration models~\cite{balakrishnan2019voxelmorph,bone2020learning} with substantially increased accuracy. The developed MetaMorph has great potential in various image-guided clinical interventions, e.g., real-time image-guided navigation systems for tumor removal surgery.


\section{Background: Diffeomorphic Image Registration}
In this section, we briefly review the concept of the diffeomorphic image registration in the setting of large deformation diffeomorphic metric mapping (LDDMM) with a geodesic shooting algorithm~\cite{beg2005computing,vialard2012diffeomorphic,miller2006geodesic}. 

Let $S$ be the source image and $T$ be the target image defined on a $d$-dimensional torus domain $\Gamma = \mathbb{R}^d / \mathbb{Z}^d$ ($S(x), T(x) : \Gamma \rightarrow \mathbb{R}$). 
The problem of diffeomorphic image registration is to find the geodesic (a.k.a. shortest path) to generate time-varying diffeomorphisms $\{\psi_t(x)\}: t \in [0,1] $ such that $S \circ \psi_1$ is similar to $T$, where $\circ$ is an interpolation operation that deforms $S$ by the smooth deformation field $\psi_1$. This is typically formulated as an optimization problem by minimizing an explicit energy function over the transformation fields $\psi_t$ as
\begin{equation}
\label{eq:imReg} 
\E(v_t) = \Dist [S \circ \psi_1(v_t), T] + \Reg[\psi_t(v_t)],
\end{equation} 
where the distance function $\Dist(\cdot , \cdot)$ measures the image dissimilarity between the source and the deformed image. Commonly used distance functions include a sum-of-squared difference
of image intensities~\cite{beg2005computing}, normalized cross correlation~\cite{avants2008symmetric}, and mutual information~\cite{wells1996multi,zhao2019region}. The regularization term $\Reg(\cdot)$ is a constraint that enforces the spatial smoothness of transformations, arising
from a distance metric on the tangent space $V$ of diffeomorphisms, i.e., an integral over the norm of time-dependent velocity fields $\{v_t(x)\} \in V$,
\begin{equation}\label{eq:distance}
    \Reg(\psi_t) = \int_0^1 (L v_t, v_t) \, dt,  \quad \text{with} \quad \frac{d\psi_t}{dt} = - D\psi_t\cdot v_t, 
\end{equation}
where $L: V\rightarrow V^{*}$ is a symmetric, positive-definite differential operator that maps a tangent vector $ v_t\in V$ into its dual space as 
a momentum vector $m_t \in V^*$. We typically write $m_t = L v_t$, or $v_t = K m_t$, with $K$ being an inverse operator of $L$. The notation $(\cdot, \cdot)$ denotes the pairing of a momentum vector with a tangent vector, which is similar to an inner product. Here, the operator $D$ denotes a Jacobian matrix and $\cdot$ represents element-wise matrix multiplication.

A geodesic curve with a fixed endpoint is characterized by an extremum of the energy function~\eqref{eq:distance}
that satisfies the Euler-Poincar\'e differential (EPDiff) equation~\cite{arnold1966geometrie,miller2006geodesic}, 
\begin{align}
    \frac{\partial v_t}{\partial t} = - K \left[(D v_t)^T \cdot m_t + D m_t \cdot v_t + m_t \cdot \operatorname{div} v_t \right],
    \label{eq:epdiff}
\end{align}
where $\operatorname{div}$ is the divergence. This process in Eq.~\eqref{eq:epdiff} is known as {\em geodesic shooting}, stating that the geodesic path $\{\psi_t\}$ can be uniquely 
determined by integrating a given initial velocity $v_0$ forward in time by using the rule~\eqref{eq:epdiff}. 

Therefore, we rewrite the optimization of Eq.~\eqref{eq:imReg} equivalently as
\begin{align}
\label{eq:imRegv0} 
\E(v_0) =\Dist [S \circ \psi_1(v_0), T] + (L v_0, v_0), \, \, \text{s.t.} \, \, \text{Eq.~\eqref{eq:distance}} \&  \text{Eq.~\eqref{eq:epdiff}} .
\end{align}


\section{Our Model: MetaMorph}
The objective function of diffeomorphic image registration in Eq.~\eqref{eq:imRegv0} works well under the condition that images are ideally of good quality with preserved topology. This assumption breaks when corruptions such as appearance changes or occlusions occur. In this section, we first define an objective function of the metamorphic image registration that considers the modeling of appearance changes. An appearance-aware regularization is developed to effectively suppress the negative influences of appearance changes in typical diffeomorphic image registration algorithms. We then develop a joint learning framework that  includes i) a segmentation network for appearance change detection, and ii) a metamorphic registration network incorporating the newly formulated objective function as part of the network loss. 

\paragraph{\bf{Appearance-aware regularization.}} The purpose of metamorphic image registration is to find an optimal transformation $\psi (v_0, \delta)$ that is composed of two variables: the optimal initial velocity field $v_0$, and the appearance change $\delta$. A recent work proposed to learn these variables via disentangled latent representations in an encoder-decoder neural network~\cite{bone2020learning}. However, it is extremely challenging for this algorithm to differentiate the variations of image intensities caused by geometric transformations from appearance changes since they unavoidably compensate for each other. The ambiguity introduced by optimizing two compensating variables without any guidance fails to search for accurate registration solutions. Additionally, this makes the algorithm highly sensitive to network parameters with an increased risk of poor convergence. To alleviate this issue, we introduce an appearance-aware regularization in the registration framework, guided by learned segmentations of the appearance-changing areas. 

Assume $U$ is a union of the learned segmentations of appearance-changing areas from the source image $S$ and the target image $T$. Analogous to Eq.~\eqref{eq:imRegv0}, we define the appearance-aware regularization $\textbf{Reg}^*(\cdot)$ in the space of initial velocity fields. To suppress the effects of appearance variations, we piecewisely constrain the initial velocity fields through a segmentation indicator, i.e.,
\begin{equation}
\label{eq:newreg}
 \textbf{Reg}^*(v_0) =  \left(L (v_0 \odot (1-U)), v_0 \odot (1-U) \right), \, \, \text{s.t.} \, \, \text{Eq.}~\eqref{eq:epdiff},
\end{equation}
where $\odot$ represents an element-wise multiplication between a vector field and a scalar field. For the purpose of notation simplicity, we define $\hat{v}_0 \overset{\Delta}{=} v_0 \odot (1-U)$ in the following sections. 

With the newly defined regularization in Eq.~\eqref{eq:newreg}, we arrive at the objective function of metamorphic image registration as
\begin{equation}
\textbf{E}^*[\hat{\psi} (\hat{v}_0)] = \textbf{Dist}^*[\hat{S} \circ \hat{\psi}_1 (\hat{v}_0), \hat{T} ] + \textbf{Reg}^* (\hat{v}_0),
\label{eq:metaenergy}
\end{equation}
 where $\hat{S}$ and $\hat{T}$ denotes the source and target images with appearance changes masked out, i.e., $\hat{S} = S \odot (1-U)$, and $\hat{T} = T \odot (1-U)$.  Here, the $\textbf{Dist}^*[\cdot, \cdot]$ is the image dissimilarity term that measures the dissimilarity between the consistent area between the deformed image and target. 

\subsection{Predictive Metamorphic Image Registration}
\begin{figure}[t]
\begin{center}
 \includegraphics[width=.93\textwidth] {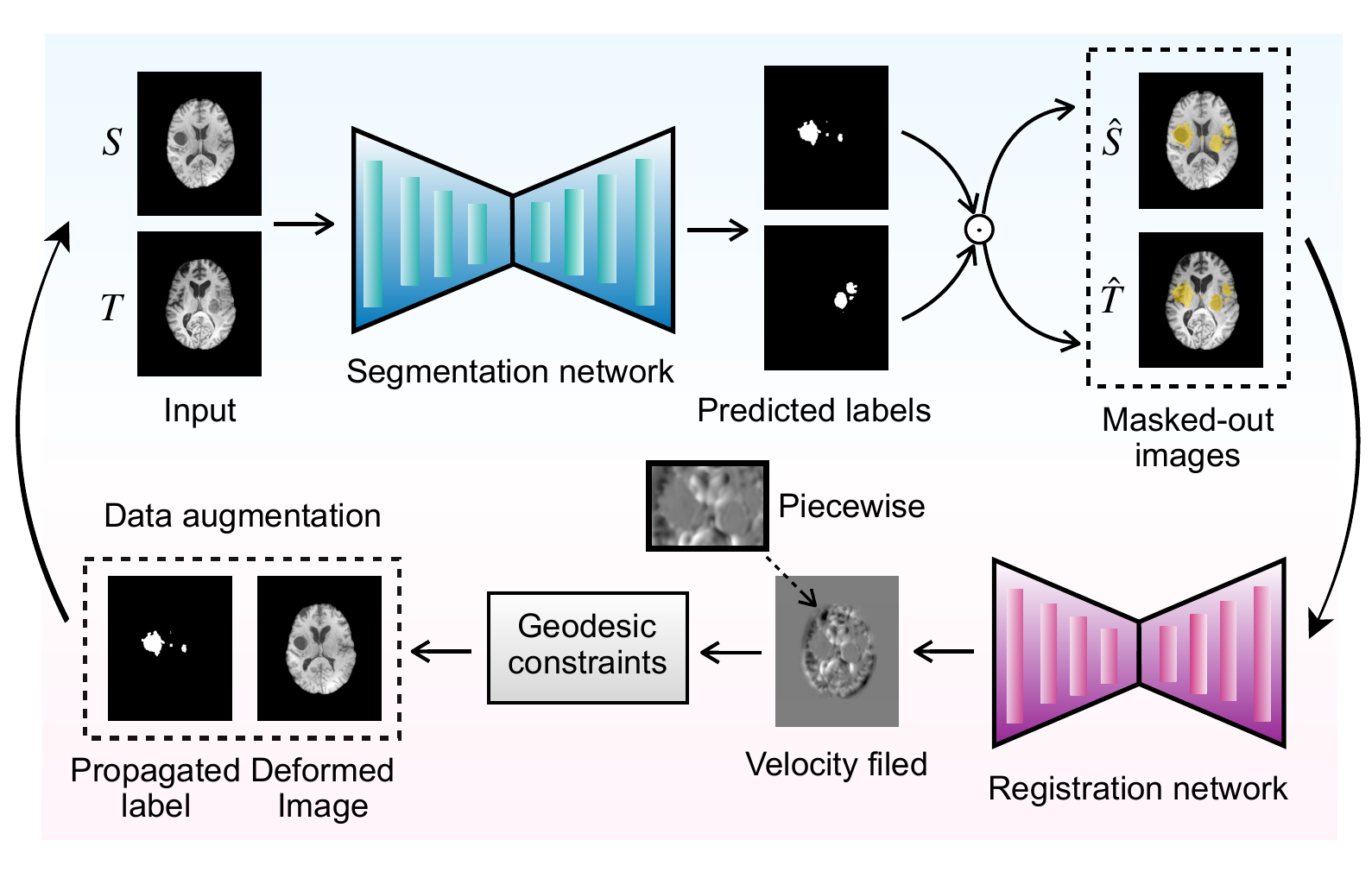}
     \caption{An illustration of the network architecture for MetaMorph. Top left to right: input a pair of images into a segmentation network, and apply predicted labels onto images to mask out the appearance change. Bottom right to left: input a pair of images (with masked-out appearance change) to the registration network and predict a piecewise velocity field, integrate geodesic constraints, and produce a deformed image and transformation-propagated segmentation. The deformed images and labels are circulated into the segmentation network as augmented data.}
\label{fig:metareg}
\end{center}             
\end{figure}

We develop a deep learning framework to jointly learn the segmentation for appearance change and the masked-out velocity field $\hat{v}_0$. An overview of our proposed MetaMorph architecture is shown in Fig.~\ref{fig:metareg}.   

Appearance change can be masked by a fixed foreground segmentation via pre-running image segmentation algorithms~\cite{niethammer2011geometric,patenaude2011bayesian}. However, performing manual annotations of segmentation labels is time and labor-consuming. In this work, instead of using a fixed mask, we treat the appearance change as a variable from the segmentation network and jointly optimize with the optimal registration solution. We utilize an encoder-decoder based neural network to learn the segmentation masks and then apply them to the associate image pairs for masking out the appearance change. Although we adopt UNet-based architecture for segmentation in this work~\cite{ronneberger2015u}, other networks such as recurrent residual neural networks~\cite{alom2018recurrent}, transformer-based networks~\cite{chen2021transunet,hatamizadeh2022unetr} can also be easily plugged into the proposed method.  

With the developed segmentation network, now we are ready to formulate the loss function of MetaMorph,
\begin{align}
\label{eq:meta_formula} 
\ell =    \textbf{Dist}^*[\hat{S} \circ \hat{\psi}_1 (\hat{v}_0), \hat{T} ] + \textbf{Reg}^* (\hat{v}_0) + \gamma \cdot \ell_{seg}  , \,\, \,  \text{s.t.} \, \, \text{Eq.~\eqref{eq:newreg}}.
\end{align}

Here, $\gamma$ is a weighting parameter that balances the segmentation and registration loss, $\ell_{seg}$ is a segmentation loss that maximizes the S{\o}rensen-Dice coefficient~\cite{dice1945measures} between ground truth $y$ and the predicted $\hat{y}$, 
\begin{align}
    \ell_{seg} &= 1-\text{Dice} (y, \hat{y} ), 
    \label{loss:seg}
\end{align}
where $\text{Dice} (y, \hat{y} ) = 2(|y| \cap |\hat{y}|)/(|y| + |\hat{y}|) $.

We adopt an approximated region-based mutual information (RMI)~\cite{zhao2019region}, which is a broadly-used distance metric for images from different domains. For simplicity, we let $\hat{S}_{\psi}$ denote the deformed image. Let $f(\hat{S}_{\psi})$ and $f(\hat{T})$ denote the probability density functions for the deformed image and target respectively, and their joint probability density function is $f (\hat{S}_{\psi}, \hat{T})$. The image dissimilarity with RMI can be formulated as
\begin{align}
\label{eq:newdissim} 
\textbf{Dist}^*[\hat{S}_{\psi},\hat{T}] &=\text{RMI} (\hat{S}_{\psi}, \hat{T}) =  \int_{\hat{S}_{\psi}}\int_{\hat{T}} f (\hat{S}_{\psi}, \hat{T}) \log \frac{f (\hat{S}_{\psi}, \hat{T})}{f(\hat{S}_{\psi})f(\hat{T})}  \nonumber
\\& \approx l_{ce} (\hat{S}_{\psi},\hat{T}) - \frac{1}{B}\sum^{B}_{b=1} I_b (\hat{T}; \hat{S}_{\psi}),
\end{align}
where $L_{ce}(\cdot, \cdot)$ is a cross entropy loss between two images. The $I_b (\cdot;\cdot)$ is a batch-wise lower bound that $I_b (\hat{T}; \hat{S}_{\psi}) = \frac{1}{2}\log [\det (\Sigma_{\hat{T}|\hat{S}_{\psi}})]$, where $\Sigma_{\hat{T}|\hat{S}_{\psi}}$ is the posterior covariance matrix of $\hat{T}$ (a symmetric positive semi-definite matrix), given $\hat{S}_{\psi}$. Here $B$ denotes the number of images in a mini-batch $b$. Please refer to \cite{zhao2019region} for more derivation details. 

We develop an alternating optimization scheme~\cite{nocedal1999numerical} to minimize the network loss defined in Eq.~\eqref{eq:meta_formula}. All network parameters are optimized jointly by alternating between the training of segmentation and image registration. A summary of our joint learning of MetaMorph is in Alg.~\ref{alg1}.
\begin{algorithm}[!h]
\SetAlgoLined
\SetArgSty{textnormal}
\SetKwInOut{Input}{Input}
\SetKwInOut{Output}{Output}
\DontPrintSemicolon
\Input{Source and target images, the number of iterations $q$. }
\Output{Segmentation labels, the deformed image, and the transformation.}
 \For{$i=1$ to $q$} {
\tcc{Train image segmentation network}
Minimize the segmentation loss in Eq.~\eqref{loss:seg};

Output the predicted segmentations and adopt both labels to mask appearance change in images;

\tcc{Train appearance-aware registration network}
Minimizing the metamorphic loss in Eq.~\eqref{eq:metaenergy} with appearance-aware geodesic constraints;

Output the predicted velocity field and the deformed image;
}
\textbf{Until} convergence
\caption{Joint learning of MetaMorph.} \label{alg1}
\end{algorithm}

\section{Experimental Evaluation}
To demonstrate the effectiveness of the proposed model, we compare both segmentation and registration tasks with state-of-the-arts.

\paragraph*{\bf Data.} For 3D  brain tumor MRI scans with tumor segmentation labels, we include $100$ public T1-weighted brain scans of different subjects from Brain Tumor Segmentation (BraTS) ~\cite{baid2021rsna,menze2014multimodal} challenge 2021. We also include 28 landmarks (16 for brain ventricle and 12 for corpus callosum) that are annotated by clinicians to better evaluate the image registration performance. All MRIs are $155 \times 240 \times 240$, $1.25mm^{3}$ isotropic voxels. As a preprocessing step, we run affine registration, intensity normalization, and bias field correction on all images. 

\paragraph*{\bf Experiments.} 
We compare our metamorphic image registration method with two registration baselines, 
an unsupervised predictive diffeomorphic registration method (VoxelMorph as VM)~\cite{balakrishnan2019voxelmorph}, and a metamorphic autoencoder (MAE)~\cite{bone2020learning} that learns disentangled appearance and shape representations. To better visualize the deformations, we show predicted transformation grids and deformed images with transformation-propagated landmarks for all methods. Quantitatively, we compute the $L_2$ distance of landmarks as registration error between the propagated and the target frames over 60 pairs.

We evaluate the brain tumor segmentation via computing Dice score~\cite{dice1945measures} by comparing MetaMorph with three segmentation backbones, U-Net architecture~\cite{ronneberger2015u}, U-Net based on recurrent residual convolutional neural network (R2-Unet)~\cite{alom2018recurrent}, and transformer-based Unet (UnetR)~\cite{hatamizadeh2022unetr}. We also show the performance of MetaMorph by replacing the segmentation module in our model with all backbones (named MetaMorph:Unet, MetaMorph:R2-Unet, and MetaMorph:UnetR). We visualize the predicted segmentations overlaid with testing images across all methods.

\paragraph*{\bf Parameter Settings.} We set parameter $\alpha=3$ for the operator $L$, the number of time steps for Euler integration in EPDiff (Eq.~\eqref{eq:epdiff}) as $10$. We set the weight parameter $\gamma = 0.5$ and the batch size as $4$. We use an adaptive cosine annealing learning rate scheduler that starts from an initial value at $\eta = 5e-4$ for network training. We run all models for $100$ epochs with Adam optimizer and save the networks with the best validation performance. The training and prediction procedure of all learning-based methods are performed on two Nvidia GTX 2070Ti GPUs. We run five-fold cross validation and split the images by using $70\%$ as training images, $20\%$ as validation images, and $10\%$ as testing images. 

\begin{figure}[!thb]
\begin{center}
 \includegraphics[width=1.0\textwidth] {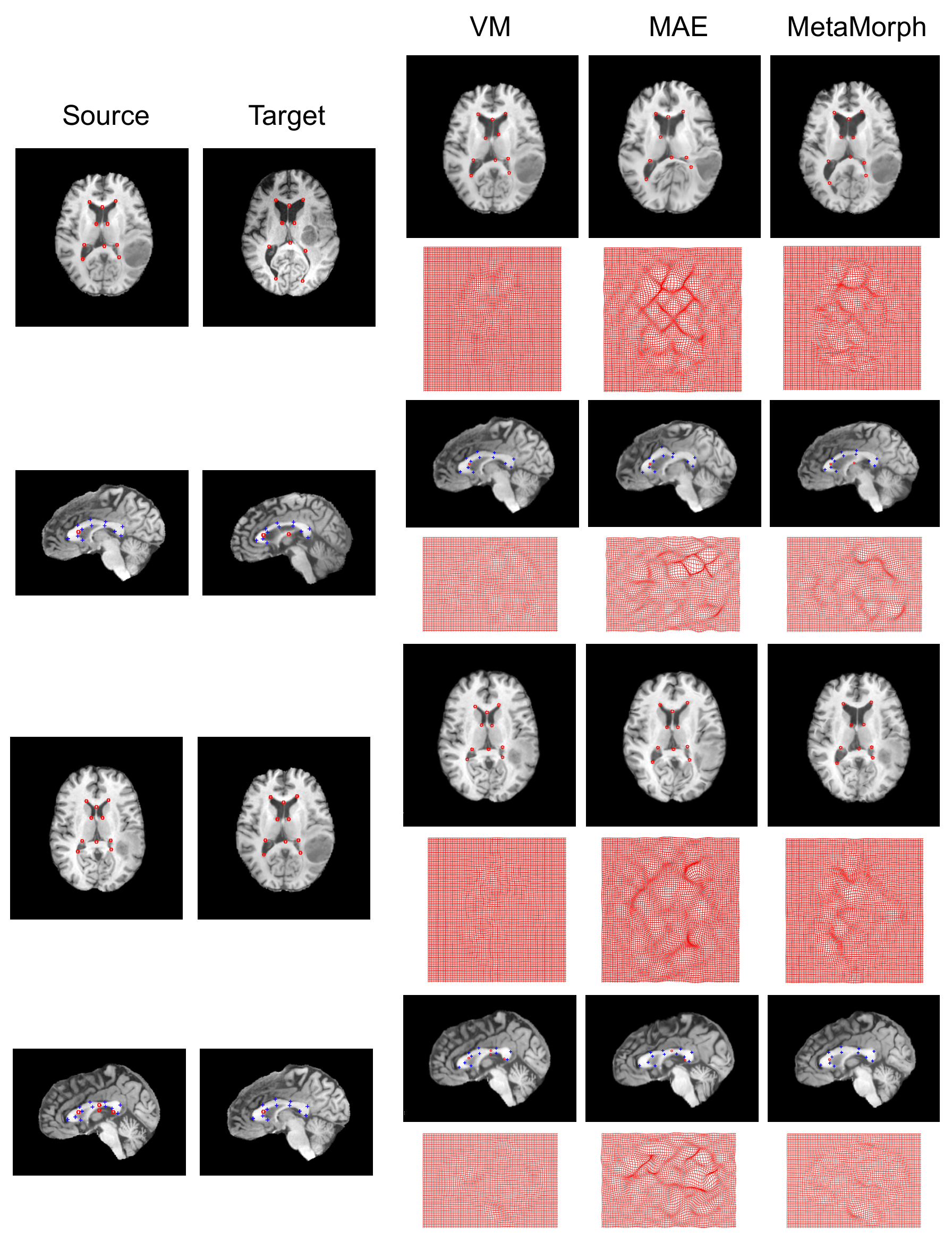}
     \caption{Image registration performance comparison for all methods. From left to right, source, target, deformed images by VoxelMorph (VM), metamorphic autoencoder (MAE), and our method. All images are overlaid with annotated landmarks (red circle for ventricle and blue cross for corpus callosum). }
\label{fig:reg}
\end{center}             
\end{figure}

\paragraph*{\bf Results.}
Fig.~\ref{fig:reg} visualizes the image registration prediction of two 3D brain MRIs of study across all methods. It shows MetaMorph significantly outperforms both VM and MAE. General diffeomorphic registration models (e.g., VM) without an appearance-control mechanism may fail and produce less satisfied deformed images without sufficient deformations. MAE offers accurate deformations to a certain level while it produces artifacts. By excluding the appearance change, MetaMorph more accurately deforms all regions (e.g., ventricles and corpus callosum). It also shows that our propagated landmarks align best with the target.

\begin{figure}[!htb]
\begin{center}
 \includegraphics[width=.98\textwidth] {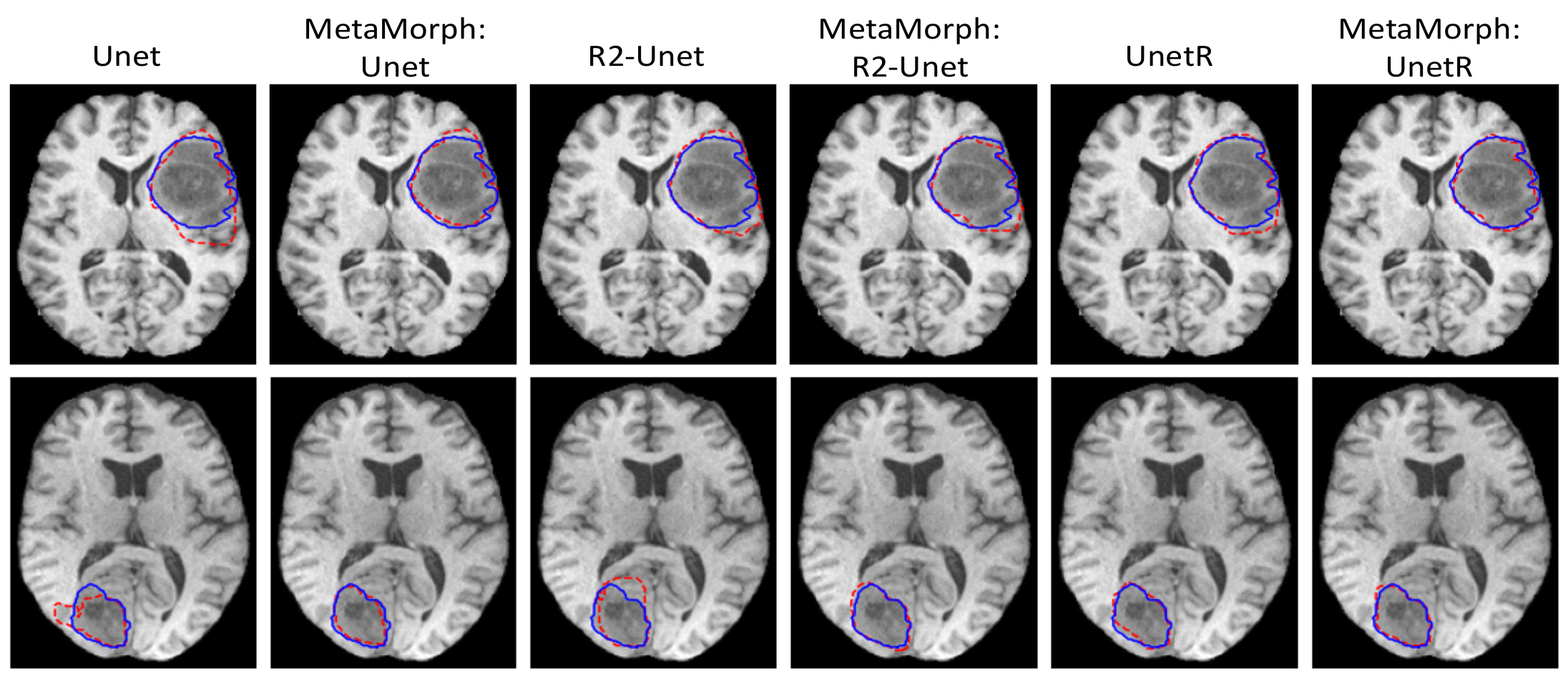}
     \caption{Image segmentation visualization for all methods. Left to right: overlaid segmentation map comparison between the predicted label (red) and the ground truth (blue) for Unet, MetaMorph: Unet, R2-Unet, MetaMorph: R2-Unet, UnetR and MetaMorph: UnetR.}
\label{fig:seg}
\end{center}             
\end{figure}

Fig.~\ref{fig:seg} shows two examples of image segmentation performance comparison for all methods. It indicates that MetaMorph-based models predict better segmentation labels (closer to ground truth) than original backbones. The predicted labels by MetaMorph have slightly better segmentations of the brain tumor boundary. This is because we use deformed images and labels that are produced by a joint registration framework as augmented data for each subject; thus learning a broader spectrum for appearance variation in data and offering more accurate prediction when new testing data arrives.

\begin{figure}[!htb]
\begin{center}
 \includegraphics[width=.98\textwidth] {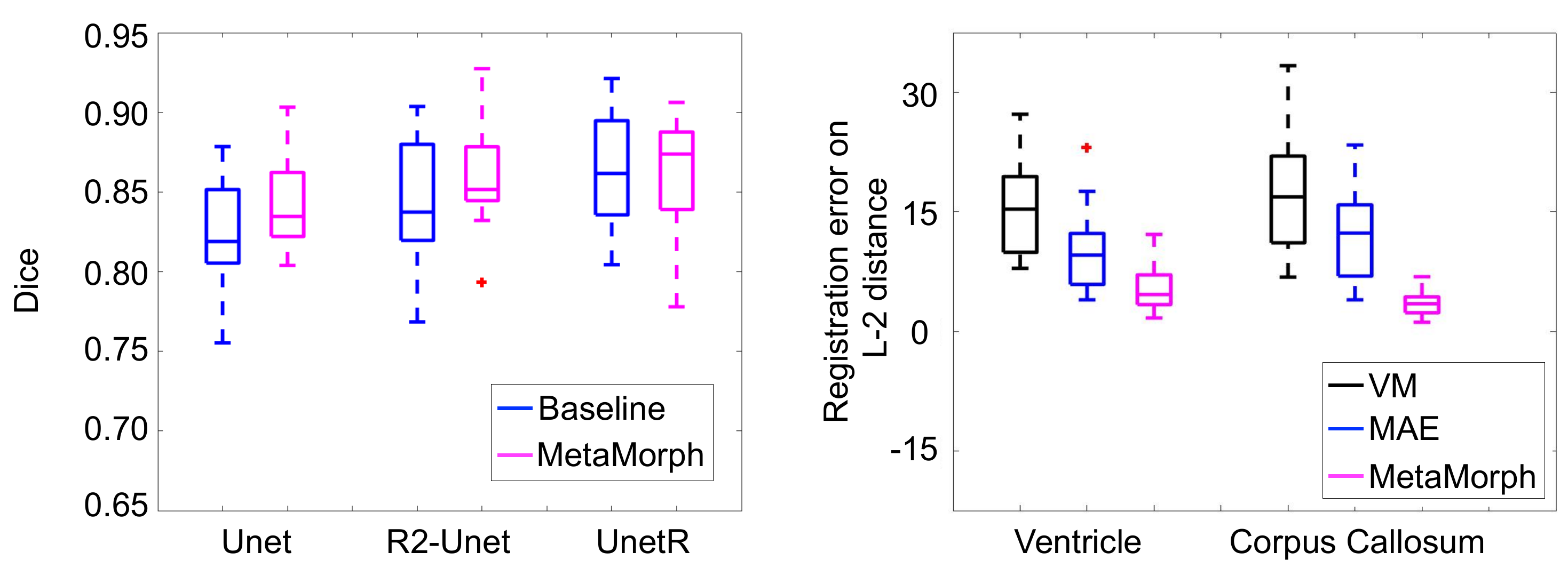}
     \caption{Left: Dice comparison on brain tumor segmentation across all methods over images. The means of baseline vs. \textbf {our method} are 0.815/\textbf{0.834}, 0.835/\textbf{0.856}, 0.861/\textbf{0.874}; Right: registration error (computed on $L_2$ distance) of two anatomical landmarks for 60 brain pairs. The means of errors for VM vs. MAE vs. \textbf {our method} are 15.02/10.53/\textbf{4.64}, 16.48/13.59/\textbf{4.10}.}
\label{fig:stat}
\end{center}             
\end{figure}

Fig.~\ref{fig:stat} (left panel) statistical reports the Dice coefficient comparison. It shows that MetaMorph consistently achieves a higher segmentation accuracy than backbones. Transformer-based methods (UnetR-based) produce the highest Dice for all methods. Fig.~\ref{fig:stat} (right panel) reports the landmark-based registration error between the target image and the deformed image. MetaMorph outperforms other methods with the lowest error, indicating our proposed method finishes the metamorphic image registration task with higher accuracy.   

\section{Conclusion}
We present a predictive metamorphic image registration model, MetaMorph, via deep neural networks in this paper. Different from existing models that have limited control over appearance change, we develop a joint learning framework that adopts a segmentation module to accurately guide the registration network to learn diffeomorphic transformation fields. The developed segmentation module maximally excludes the disadvantageous effect caused by appearance change for learned deformations; thus enabling more precise correspondence alignment between deformed and target frames. Experimental results on 3D brain MRIs with real tumors show that our proposed framework yields a better registration as well as a segmentation model. While our algorithm is presented in the setting of LDDMM with geodesic shooting, the theoretical development is generic to other deformation models, e.g., stationary velocity fields~\cite{arsigny2006log}. Our model has great clinical potential on solving one of the most challenging registration problems, e.g., real-time brain shift estimation between preoperative and intraoperative MRI scans with missing data values. Interesting future works of MetaMorph will be i) building a probabilistic model to quantify the registration uncertainty along the boundary of tumor areas and ii) extending the proposed method to more advanced clinical scenarios that appearance changes are difficult to detect, e.g., real-time automated image registration for ultrasound images.
\paragraph{\bf Acknowledgments} This work was supported by NSF CAREER Grant 2239977.
\bibliographystyle{splncs04}
\bibliography{IPMI2023_ref}
\end{document}